\def\year{2019}\relax

\documentclass[letterpaper]{article}  
\usepackage{aaai19}  
\usepackage{times}  
\usepackage{helvet}  
\usepackage{courier}  
\usepackage{url}  
\usepackage{graphicx}  

\usepackage{amsmath}
\usepackage{booktabs}
\usepackage{multirow}
\usepackage{subfigure}

\begin{document}
%
\title{Recurrent Skipping Networks for Entity Alignment}
\author{Lingbing Guo, Zequn Sun, Ermei Cao, Wei Hu\\
	State Key Laboratory for Novel Software Technology,\\ Nanjing University, China\\
	Email: lbguo.nju@gmail.com, zqsun.nju@gmail.com, emcao.nju@gmail.com, whu@nju.edu.cn
}
\maketitle


\begin{abstract}
We consider the problem of learning knowledge graph (KG) embeddings for entity alignment (EA). Current methods  use the embedding models mainly focusing on triple-level learning, which lacks the ability of capturing long-term dependencies existing in KGs. Consequently, the embedding-based EA methods heavily rely on the amount of prior (known) alignment, due to the identity information in the prior alignment cannot be efficiently propagated from one KG to another. In this paper, we propose RSN4EA (recurrent skipping networks for EA), which leverages biased random walk sampling for generating long paths across KGs and models the paths with a novel recurrent skipping network (RSN). RSN integrates the conventional recurrent neural network (RNN) with residual learning and can largely improve the convergence speed and performance with only a few more parameters. We evaluated RSN4EA on a series of datasets constructed from real-world KGs. Our experimental results showed that it outperformed a number of state-of-the-art embedding-based EA methods and also achieved comparable performance for KG completion.
\end{abstract}


\section{Introduction}
\label{sect:intro}

Knowledge graphs (KGs) have become one of the most important resources for many areas, e.g., question answering and recommendation. Many KGs are created and maintained by different parties and in various languages, which makes them inevitably heterogeneous. Entity alignment (EA) aims to address this problem. It finds entities in two KGs referring to the same real-world object.

Recently, a number of methods start to consider leveraging the representation learning techniques for EA \cite{MTransE,JAPE,BootEA,KDCoE}. Most of them are based on a classical KG embedding model called TransE \cite{TransE}, which interprets each triple $(s,l,o)$ in a KG as $s+l\approx o$, where $s$ and $o$ denote the subject and object entities respectively, and $l$ denotes the relation label between them. However, these methods may suffer from the problem of modeling multi-relational triples \cite{TransR}. Moreover, they only concern triple-level embeddings, i.e., they train a triple $(s,l,o)$ only using the embeddings of $s, l$ and $o$. Although the information of multi-hop neighbors can be passed during several rounds of mini-batches using back propagation \cite{Survey}, the efficiency would be severely affected, especially for the case of crossing KGs. A path-based method IPTransE \cite{IPTransE} tries to learn inferences among relations, but it still concentrates on the triple-level embedding learning. The long-term dependencies of entities are ignored by the current methods. For EA, the triple-level embedding learning limits the identity information propagating across KGs, especially for the entities which are not well connected with other entities or far away from the entities in prior alignment (i.e., entity alignment known ahead of time). Also, the triple-level learning only uses triples involved in prior alignment to deliver information across KGs, it also makes the current methods heavily rely on the amount of prior alignment. 

KGs can be regarded as multi-relational graphs and triples are just paths of length 1. If a KG embedding model is capable of being aware of the associations among entities in long paths, the trained embeddings would contain much richer information and thus help EA. However, none of the current EA methods takes modeling KG paths into consideration. To model KG paths, there exist two challenges that need to be solved. The first one is how to obtain these paths. A KG may have millions (even billions) of triples and the number of its paths is also huge. It is difficult, if not impossible, to use all of them for training. The second challenge is how to model these paths. The edges in the paths have labels and directions. We cannot simply ignore them when modeling the dependencies among entities. 

In this paper, we propose a new method, called RSN4EA (recurrent skipping networks for EA), which employs random walk sampling to efficiently sample paths across KGs, and models the paths with a novel recurrent skipping network (RSN). According to the network representation learning \cite{DeepWalk,node2vec}, an appropriate sampling method reduces computational complexity and often brings good performance. So, sampling paths from KGs is also worth exploring. Compared with networks, which typically consider edges with no labels or directions, KGs have more complex graph structures. Furthermore, our problem requires to propagate the identity information through the paths across KGs. To deal with these issues, we design a biased random walk sampling method to fluently control the depth and cross-KG biases of generated paths.

To model paths or sentences, Skip-gram \cite{word2vec} is widely used in the natural language processing area. It can efficiently encode the neighboring information into embeddings, which is important for discovering clusters or communities of related nodes (words). However, Skip-gram does not consider the order of nodes, while relations in KGs have different directions and enormous labels. The recurrent neural network (RNN) is a popular sequential model. It assumes that the next element only depends on the current input and the previous hidden state. But this assumption has inconsiderations for KG path modeling. Take a path $(s,l,o),(o,l',o')$ for example, RNN uses the input $l'$ and the previous hidden state $h_{o}$ to infer $l'\rightarrow o'$. However, all the context of $l'$ is mixed in $h_{o}$, which overlooks the importance of $o$. Note that this path is also constituted by two triples. To predict the object entity of $(o, l', ?)$, both $o$ and $l'$ should be more appreciated than others. To achieve this, we combine the idea of residual learning \cite{ResNet} with RNN to let the output hidden state of $l'$ learn a residual between the subject $o$ and the desired prediction $o'$, which leads to our recurrent skipping network (RSN).

To evaluate RSN4EA, we built a series of datasets from real-world KGs. The previous work did not carefully consider the density and degree distributions of their datasets, which makes the datasets used in their experiments much denser than the original KGs. Also, their sampling methods are vague. In this paper, we created four couples of datasets, which were sampled with a reliable method and consider mono/cross-lingual scenarios and normal/high density.

The main contributions of this paper are listed below: 
\begin{itemize}
	\item We propose RSN4EA, an end-to-end framework for EA, which is capable of capturing long-term dependencies existing in KGs. 
	\item We design a biased random walk sampling method specific to EA, which generates desired paths with controllable depth and cross-KG biases. 
	\item To revise the inconsideration of RNN for KG path modeling, we present RSN, which leverages the idea of residual learning and can largely improve the convergence speed and performance. 
	\item To demonstrate the feasibility of our method, we carried out EA experiments on the datasets with different density and languages. The results showed that our method stably outperformed the existing methods. Also, RSN4EA achieved comparable performance for KG completion.
\end{itemize}


\section{Related Work}
\label{sect:work}

We divide the related work into three areas: KG representation learning, embedding-based EA and network representation learning. We discuss them in the rest of this section.

\subsection{KG Representation Learning}
KG representation learning has been widely studied in recent years \cite{Survey}. One of the most famous translational methods is TransE \cite{TransE}, which models a triple $(s,l,o)$ as $s+l\approx o$. TransE works well for one-to-one relationships, but fails to model more complex relationships like one-to-many and many-to-many. TransR \cite{TransR} tries to solve this problem by involving a relation-specific matrix $\mathbf{W}_l$ to project $s,o$ by $\mathbf{W}_l$. PTransE \cite{PTransE} leverages path information to learn inferences among relations. For example, if there exist two triples $(e_1, l_1, e_2), (e_2, l_2, e_3)$, which form a path in KG, and another triple $(e_1, l_x, e_3)$ holds simultaneously, PTransE models the path information by learning $l_1 \oplus l_2 \approx l_x$, where $\oplus$ denotes the operator used to merge $l_1, l_2$. KG completion is the most prevalent task for KG representation learning, and there also exist some non-translation methods that are particularly tailored for KG completion \cite{ComplEx,ConvE}.

\subsection{Embedding-based Entity Alignment}

Existing embedding-based EA methods are usually based on TransE. Specifically, MTransE \cite{MTransE} separately trains the entity embeddings of two KGs and learns various transformations to align the embeddings. JAPE \cite{JAPE} is also based on TransE but learns the embeddings of two KGs in a unified space. Additionally, JAPE leverages attributes to refine entity embeddings. IPTransE \cite{IPTransE} employs an iterative process on the original PTransE \cite{PTransE} for EA. Different from our method, it still concentrates on triple-level learning and does not consider the dependencies among entities in KG paths. BootEA \cite{BootEA} takes bootstrapping into consideration and uses a sophisticated strategy to update alignment during iterations. KDCoE \cite{KDCoE} leverages co-training for separately training entity relations and entity descriptions. Like bootstrapping, propagating alignment to each other may involve errors. Moreover, it requires extra resources like pre-trained multi-lingual word embeddings and descriptions. 

Because all the aforementioned methods use TransE-like models as the basic model, they are not capable of capturing long-term dependencies in KGs and the identity information propagating between different KGs is also limited. 

\subsection{Network Representation Learning}

DeepWalk \cite{DeepWalk} is one of the most well-known models in the network representation learning area. It uses uniform random walks to sample paths in a network, and applies Skip-Gram \cite{word2vec} to model the generated paths. Skip-Gram learns the embedding of a node by maximizing the probabilities of its neighbors, which captures the information among the nodes. node2vec \cite{node2vec} proposes biased random walks to refine the process of sampling paths from a network. It smoothly controls the node selection strategy to make the random walks explore neighbors in a breadth-first-search as well as a depth-first-search fashion. In this paper, the proposed EA-specific random walk sampling is inspired by node2vec, but concentrates on generating long and cross-KG paths.

The methods in the network representation learning area mainly focus on discovering clusters or communities of related nodes. However, they are inappropriate to EA, since EA requires identifying entity alignment in two KGs.


\section{Method Overview}


A KG is defined as a directed multi-relational graph whose nodes correspond to entities and edges are of the form $(subject,label,object)$ (denoted as $(s,l,o)$), each of which indicates that there exists a relation of name $label$ between the entities $subject$ and $object$. 

EA is the task of finding entities in two KGs that refer to the same real-world object. In many cases (e.g., Linked Open Data), a subset of aligned entities, called prior alignment, is known as training data. Based on it, many existing methods, such as \cite{IPTransE,JAPE,BootEA}, merge the two KGs into a connected joint graph and learn entity embeddings on it.

\begin{figure}
	\centering
	\includegraphics[width=\linewidth]{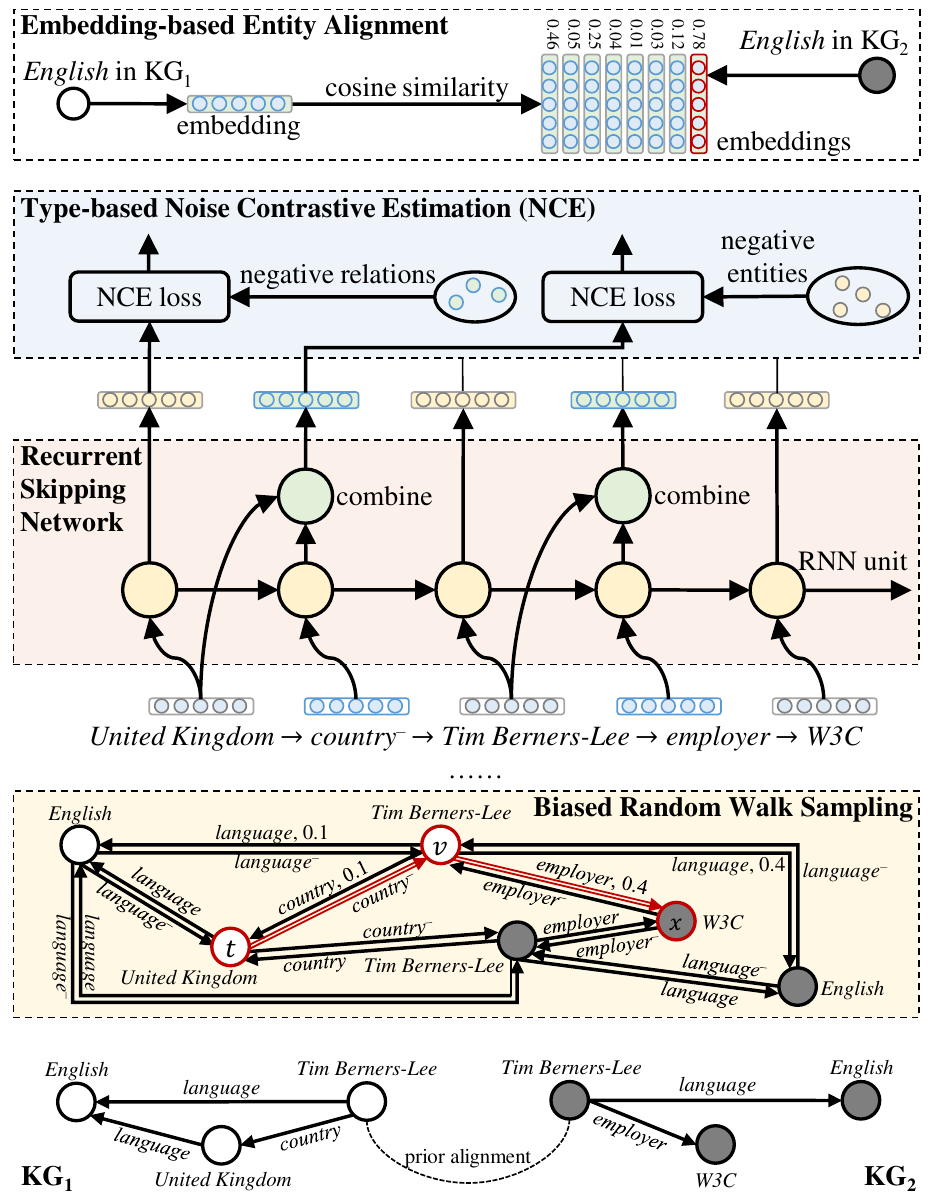}
	\caption{Architecture of the proposed method}
	\label{fig:arch}
\end{figure}

Figure~\ref{fig:arch} illustrates the architecture of our method, which accepts two KGs as input and adopts an end-to-end framework for aligning the entities between them. The main modules in the framework are described as follows:

\begin{itemize}
	\item \textbf{Biased random walk sampling.} To leverage graph sampling for EA, we first create a joint graph between the two KGs by copying the edges of one entity in prior alignment to another. Additionally, since the relation directions between entities are often arbitrary, we add a virtual reverse relation, marked by ``$^-$", for each existing relation. Thus, the object entity in a triple can follow the reverse relation to reach the subject entity. Figure~\ref{fig:arch} exemplifies the joint graph of KG$_1$ and KG$_2$ with reverse relations.
	
	\ \ \ \ Then, we conduct the biased random walk sampling on the joint graph to explore longer and cross-KG paths. We describe the details in the next section. Finally, each path, e.g., $(e_1,l_1,e_2),(e_2,l_2,e_3),\ldots,(e_{T-1},l_T,e_T)$, is converted into a KG sequence $e_1\rightarrow l_1\rightarrow e_2\rightarrow\cdots\rightarrow e_{T-1}\rightarrow l_T\rightarrow e_T$ and fed to the next module.
	
	\item \textbf{Recurrent skipping network (RSN).} RNN is natural and flexible to process sequential data types. However, it is not aware of different element types (``entity" vs. ``relation") in KG sequences and basic KG structural units (i.e., triples). To cope with these issues, we propose RSN, which distinguishes entities from relations, and leverages the idea of residual learning by letting a subject entity skip its connection to directly participate in the object entity prediction. We present RSN in detail shortly. Each output of RSN is passed to the type-based noise contrastive estimation (NCE) for learning to predict the next element. 
	
	\item \textbf{Type-based noise contrastive estimation.} NCE \cite{NCE} is a very popular estimation method in natural language processing, which samples a small number of negative classes to approximate the integral distribution. As aforementioned, entities and relations are of different types. So, we design a type-based method to sample negative examples according to element types, and use different weight matrices and biases to respectively calculate the logits for the two types of elements. By back propagation, the embedding of each input element is not only learned from predicting its next, but associated with the elements along the KG sequence.
	
	\item \textbf{Embedding-based EA.}
	With entity embeddings from the two KGs learned in a unified space, given a source entity, its aligned target entity can be discovered by searching the nearest neighbors in this space using the cosine similarity. 
	
\end{itemize}

\section{Biased Random Walk Sampling for EA}
\label{sect:rw}

Random walks have been used as the sampling methods in network representation learning for a long time \cite{DeepWalk}. KGs share a lot of features with networks, such as large scale and sparsity. In this section, we present a biased random walk sampling method specific to EA, which can efficiently explore long and cross-KG sequences.

\subsection{Random Walk Sampling}

Given a start entity $u$ in the joint graph, an unbiased random walk method obtains the probability distribution of next entities by the following equation:
\begin{align}
\label{eq:rw}
P(c_{i+1}=x\,|\,c_{i}=v) = 
\begin{cases}
\frac{\pi_{vx}}{Z} & \text{if edge $(v,l_?,x)$ exists}\\
0 & \text{otherwise}
\end{cases},
\end{align}
where $c_i$ denotes the $i^\textrm{th}$ node in this walk and we have $c_0=u$. $l_?$ denotes an arbitrary relation from current entity $v$ to next entity $x$. $\pi_{vx}$ is the unnormalized transition probability between $v$ and $x$. $Z$ is the normalizing constant.

\subsection{Biased Random Walk Sampling}

The above random walk method decides next entities in a uniform distribution. To model KGs, the basic training unit is triple, which means that the information of near entities can be updated via back propagation in different mini-batches. However, delivering the information of farther entities only with triples is hard and low-effective. Capturing longer paths of KGs becomes helpful.

To achieve this, we employ a 2$^\textrm{nd}$-order random walk sampling method in \cite{node2vec} and propose a depth bias to smoothly control the depths of sampled paths. Formally, given an entity $v$ in the joint graph, the depth bias between $v$'s previous entity $t$ and next entity $x$, denoted by $b_{dpt}(t,x)$, is defined as follows: 
\begin{align}
\label{eq:bias_dpt}
b_{dpt}(t,x) = 
\begin{cases}
\alpha & dist(t,x) = 2\\
1-\alpha & dist(t,x) < 2
\end{cases},
\end{align}
where $dist(\cdot,\cdot)$ calculates the shortest path distance and its value must be one of $\{0,1,2\}$. Hyper-parameter $\alpha\in(0,1)$ controls the depths of random walks. To favor longer paths, we let $\alpha>0.5$. For multi-edges, we treat their  biases equal. 

Let us see Figure~\ref{fig:arch} for example. Consider a random walk that just traversed edge $(t, country^-, v)$ and now resides at $v$. The walk now needs to decide on the next step so it evaluates the transition probabilities $\pi_{vx}$ on edges $(v,l_?,x)$ leading from $v$. We set the unnormalized transition probability to $\pi_{vx}=b_{dpt}(t,x)\times w_{vx}$, where $w_{vx}$ is the static edge weight. In the case of unweighted graphs, $w_{vx}=1$. 

Furthermore, specific to EA, we propose a cross-KG bias to favor paths connecting two KGs. Formally, given an entity $v$ in the joint graph, the cross-KG bias between $v$'s previous entity $t$ and next entity $x$, denoted by $b_{crs}(t,x)$, is defined as follows: 
\begin{align}
\label{eq:bias_crs}
b_{crs}(t,x) = 
\begin{cases}
\beta & \text{$t,x$ belong to different KGs}\\
1-\beta & \text{otherwise}
\end{cases},
\end{align}
where $\beta\in(0,1)$ is a hyper-parameter controlling the preferences of random walks across two KGs. To favor cross-KG paths, we let $\beta>0.5$. Similar to the depth bias, using previous and next entities avoids walking back and forth between only two entities in different KGs.

Finally, we combine $b_{dpt}(t,x)$ and $b_{crs}(t,x)$ into overall bias $b(t,x)$ and perform random walk sampling based on it: 
\begin{align}
\label{eq:bias}
b(t,x)=b_{dpt}(t,x)\times b_{crs}(t,x).
\end{align}

Recall the above example. According to the overall bias, the walk at $v$ prefers $W3C$ and $English$ in KG$_2$ to $English$ in KG$_1$. A KG sequence converted from this walk would be $United~Kingdom \rightarrow country^- \rightarrow Tim~Berners\text{-}Lee \rightarrow employer \rightarrow W3C$.


\section{Recurrent Skipping Networks}
\label{sect:rsn}

In this section, we first describe the conventional RNN. Then, we propose our RSN and discuss its characteristics.

\subsection{Recurrent Neural Networks}

RNN is a popular class of artificial neural network which performs well on sequential data types. Given a KG sequence $x_1 \rightarrow x_2 \rightarrow\ldots\rightarrow x_T$ as input, an RNN recurrently processes it with the following equation:
\begin{align}
\label{eq:rnn}
h_t = \tanh(\mathbf{W}_h h_{t-1} + \mathbf{W}_x x_t + b),
\end{align}
where $h_t$ is the output hidden state at time step $t$. $\mathbf{W}_h,\mathbf{W}_x$ are the weight matrices. $b$ is the bias.

RNN is capable of using a few parameters to cope with input of any length. It has achieved state-of-the-art performance in many areas. However, there still exist a few limitations when RNN is used to process KG sequences. 

First, the elements in a KG sequence are of two different types, namely ``entity" and ``relation", which always appear in an alternant order. However, the conventional RNN regards them as the same type elements like words or nodes, which makes the procedure of capturing the information in the KG sequences less effective. 

Second, any KG sequences are constituted by triples, but these basic structural units are overlooked by RNN. Specifically, let $x_t$ denote a relation in a KG sequence and $(x_{t-1}, x_t, x_{t+1})$ denote a triple involving $x_t$. As shown in Eq. (\ref{eq:rnn}), to predict $x_{t+1}$, RNN would combine the hidden state $h_{t-1}$ and the current input $x_t$, where $h_{t-1}$ is a mix of the information of all the previous elements $x_1,\ldots,x_{t-1}$. However, it is expected that the information of $x_{t-1},x_t$ in the triple can be more appreciated.

\subsection{Improving RNN with the Skipping Mechanism}

To better model KG sequences and remedy the semantic inconsideration of the conventional RNN, we propose the recurrent skipping network (RSN), which refines RNN with a simple but effective skipping mechanism. 

The basic idea of RSN is to shortcut current input entity to let it directly participate in predicting its object entity. In other words, an input element in a KG sequence whose type is ``entity" can not only contribute to predicting its next relation, but also straightly take part in predicting its object entity. Figure~\ref{fig:arch} shows an RSN example.

Formally, given a KG sequence $x_1 \rightarrow x_2 \rightarrow\ldots\rightarrow x_T$ as input, the skipping operation for an RSN is formulated as follows:
\begin{align}
\label{eq:rsn}
h'_t = 
\begin{cases}
h_t & \text{if $x_t$ is an entity}\\
\mathbf{S}_h h_t + x_{t-1} & \text{if $x_t$ is a relation}
\end{cases},
\end{align}
where $h'_t$ denotes the output hidden state of the RSN at time step $t$, and $h_t$ denotes the corresponding RNN output. $\mathbf{S}_h$ is the weight matrix. In this paper, we select weighted sum for the skipping operation, but other combination methods can be supported as well.

\paragraph{Explanation of RSN.}

Intuitively, RSN explicitly distinguishes entities and relations, and allows subject entities to skip their connections for directly participating in object entity predication. Behind this simple skipping operation, there exists a deeper explanation called residual learning.

Let $F(x)$ be an original mapping, where $x$ denotes the input, and $H(x)$ be the expected mapping. Compared to directly optimizing $F(x)$ to fit $H(x)$, residual learning hypothesizes that it is easier to optimize $F(x)$ to fit the residual part $H(x)-x$. For an extreme case, if an identity mapping is optimal (i.e., $H(x)=x$), pushing the residual to zero would be much easier than fitting an identity mapping by a stack of nonlinear layers \cite{ResNet}.

Different from ResNet \cite{ResNet} or recurrent residual network (RRN) \cite{RRN}, which were proposed to help train very deep networks, RSN employs residual learning on ``shallow" networks. The skipping connections do not link the previous input to the very deep layers, but only concentrate on each triple in a KG sequence. 

Specifically, given a KG sequence $\cdots\rightarrow x_{t-1}\rightarrow x_{t}\rightarrow x_{t+1}\rightarrow \cdots$, where $(x_{t-1}, x_{t}, x_{t+1})$ forms a triple, RRN leverages residual learning by regarding the process at each time step as a mini-residual network with the previous hidden state of RNN as input. Take time step $t$ for example, RRN regards $h_{t-1}$ as input, and learns the residual $h_t := H(h_{t-1}, x_t) - h_{t-1}$, where $H(h_{t-1}, x_t)$ denotes the expected mapping for $(h_{t-1},x_t)$. It still ignores the structure of KGs that $x_{t-1}, x_{t}$ should be more appreciated for predicting $x_{t+1}$. 

Differently, RSN leverages the residual learning in a new way. Instead of using an input as subtrahend ($h_{t-1}$), it directly chooses the subject entity $x_{t-1}$ as subtrahend. Making the output hidden state $h_t$ to fit $x_{t+1}$ may be hard, but learning the residual of $x_{t+1}$ and $x_{t-1}$ may be easier, which is the key characteristic of RSN.


\section{Experiments and Results}
\label{sect:exp}

We evaluated RSN4EA for EA using a variety of real-world datasets. In this section, we report the results compared with several state-of-the-art embedding-based EA methods. Since RSN4EA is capable of learning KG embeddings, we also conducted experiments to assess its performance on KG completion \cite{TransE}, which is a classical task for KG representation learning. 


\subsection{Datasets}

\begin{table}
	\setlength{\abovecaptionskip}{0pt}
	\setlength{\belowcaptionskip}{0pt}
	\centering
	\caption{Statistics of the datasets}
	\label{tab:dataset}
	{\scriptsize \begin{tabular}{llcccc}
			\toprule
			\multirow{2}{*}{Datasets} & \multirow{2}{*}{Sources} & \multicolumn{2}{c}{Normal} &\multicolumn{2}{c}{Dense} \\
			\cmidrule(lr){3-4}\cmidrule(lr){5-6}
			&	& \ \#Rel. & \#Rel tr. &  \#Rel. & \#Rel tr. \\ 
			\midrule
			\multirow{2}{*}{DBP-WD}  
			& DBpedia (English)   & 248   & 38,256    & 219 & 67,954 \\
			& Wikidata (English) \ & 148   & 39,605    & 137 & 76,034 \\
			\midrule 
			
			\multirow{2}{*}{DBP-YG}  
			& DBpedia (English)  & 219 & 33,571 & 206 & 71,257 \\
			& YAGO3 (English)   & 30  & 34,660 & 30  & 97,131 \\
			\midrule 
			\multirow{2}{*}{EN-FR}    
			& DBpedia (English)  & 230 &	35,139 & 218 & 71,587 \\
			& DBpedia (French)   & 181 & 32,827 & 171 & 66,283 \\
			\midrule 
			
			\multirow{2}{*}{EN-DE}    
			& DBpedia (English)  & 225 & 38,281 & 207 & 56,983 \\
			& DBpedia (German)   & 118 & 37,069 & 117 & 59,848 \\
			\bottomrule
			\multicolumn{6}{l}{We also extracted attribute triple of the sampled entities from original KGs. }\\
			
	\end{tabular}}
\end{table}

Although the datasets used by existing methods \cite{MTransE,JAPE,BootEA} are all sampled from real-world KGs, such as DBpedia and Wikidata, their density and degree distributions are quite different from the original ones. We argue that this status may prevent us from a comprehensive and accurate understanding of embedding-based EA. In this paper, we propose a segment-based random PageRank (SRP) sampling method, which can fluently control the density of sampled datasets. 

Random PageRank sampling is an efficient algorithm for large graph sampling \cite{sampling}. It samples nodes according to the PageRank weights and can assign higher biases to more valuable entities. However, due to the characteristic of PageRank, it also favors high-degree nodes. To fulfill our requirements on KG sampling, we divided the entities in a KG into segments according to their degrees and performed sampling separately. To guarantee the distributions of sampled datasets following the original KGs, we used Kolmogorov-Smirnov (K-S) test to measure the difference. We set our expectation to $\epsilon=5\%$ for all the datasets.

\begin{figure} 
\centering    
\subfigure[DBpedia]{\includegraphics[width=0.49\columnwidth]{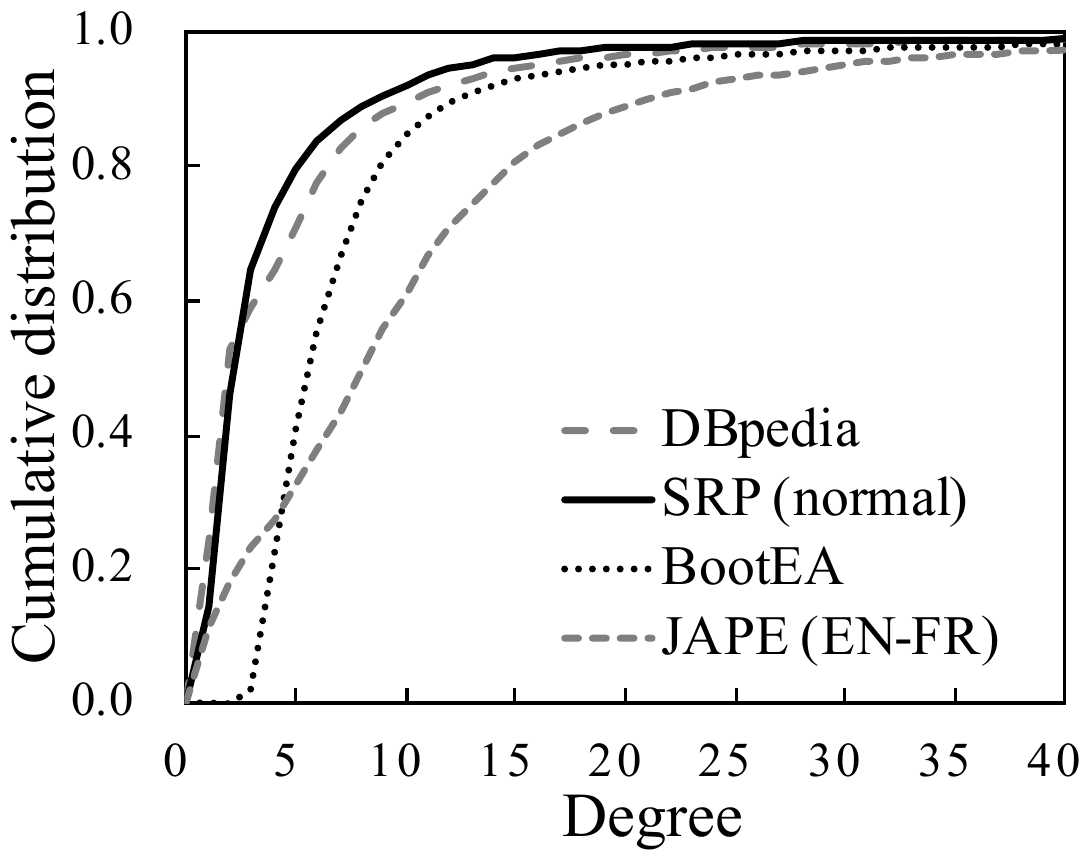}}
\subfigure[Wikidata] {\includegraphics[width=0.49\columnwidth]{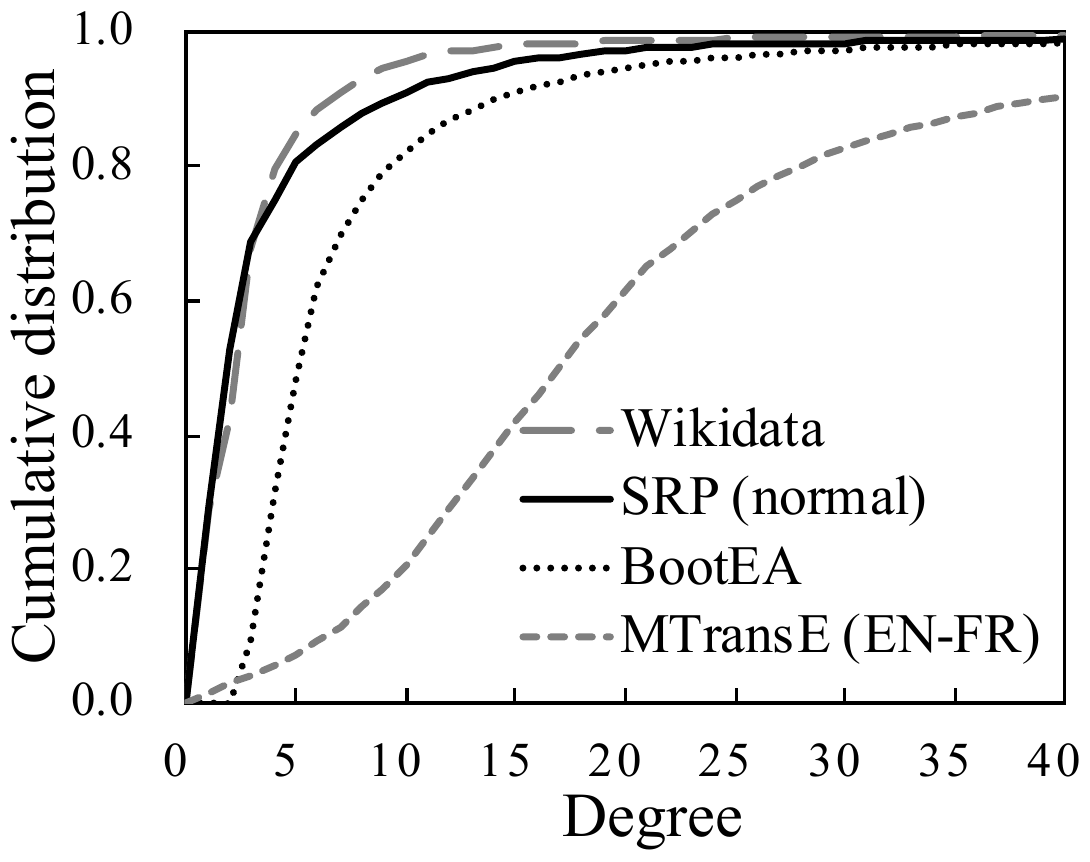}}     
\caption{Degree distributions of the datasets extracted by different methods}     
\label{fig:degree}     
\end{figure}

Based on the above sampling method, we obtained four couples of datasets to evaluate the performance of the embedding-based EA methods. The detailed statistics are shown in Table \ref{tab:dataset}. Each dataset contains nearly 15,000 entities. For the normal datasets, they follow the density of the original KGs. For the dense datasets, we randomly deleted entities with low degrees in the original KGs to make the average degree doubled, and then conducted sampling. Therefore, the dense datasets are more similar to the datasets used by the existing methods \cite{MTransE,JAPE,BootEA}. Figure~\ref{fig:degree} shows the degree distributions of source KGs and the sampled datasets from different methods. We can see that our normal datasets best represent the original KGs.

\begin{table*}
	\centering
	\caption{Entity alignment results on monolingual datasets}
	\label{tab:mono_results}
	{\scriptsize
		\begin{tabular}{lcccccc cccccc}
			\toprule
			\multirow{2}{*}{Methods} & \multicolumn{3}{c}{DBP-WD (normal)} & \multicolumn{3}{c}{DBP-WD (dense)} & \multicolumn{3}{c}{DBP-YG (normal)} & \multicolumn{3}{c}{DBP-YG (dense)} \\
			\cmidrule(lr){2-4} \cmidrule(lr){5-7} \cmidrule(lr){8-10} \cmidrule(lr){11-13} & Hits@1 & Hits@10 & MRR & Hits@1 & Hits@10 & MRR & Hits@1 & Hits@10 & MRR & Hits@1 & Hits@10 & MRR \\ 
			\midrule MTransE  & 22.3 & 50.1 & 0.32 & 38.9 & 68.7 & 0.49 & 24.6 & 54.0 & 0.34 & 22.8 & 51.3 & 0.32 \\
			IPTransE & 23.1 & 51.7 & 0.33 & 43.5 & 74.5 & 0.54 & 22.7 & 50.0 & 0.32 & 23.6 & 51.3 & 0.33 \\
			JAPE     & 21.9 & 50.1 & 0.31 & 39.3 & 70.5 & 0.50 & 23.3 & 52.7 & 0.33 & 26.8 & 57.3 & 0.37 \\
			KDCoE    & 24.6 & 51.5 & 0.34 & 56.5 & 83.1 & 0.65 & 22.7 & 47.0 & 0.31 & 56.8 & 80.4 & 0.64 \\ 
			BootEA   & 32.3 & 63.1 & 0.42 & 67.8 & 91.2 & 0.76 & 31.3 & 62.5 & 0.42 & 68.2 & 89.8 & 0.76 \\ 
			\midrule RSN4EA   & \textbf{38.8} & \textbf{65.7} & \textbf{0.49} & \textbf{76.3} & \textbf{92.4} & \textbf{0.83} & \textbf{40.0} & \textbf{67.5} & \textbf{0.50} & \textbf{82.6} & \textbf{95.8} & \textbf{0.87} \\ 
			\bottomrule
			\multicolumn{13}{l}{The best results are marked in bold. The same to the following. }\\
			
	\end{tabular}}
\end{table*}

\begin{table*}
	\centering
	\caption{Entity alignment results on cross-lingual datasets}
	\label{tab:cross_results}
	{\scriptsize
		\begin{tabular}{lcccccc cccccc}
			\toprule \multirow{2}{*}{Methods} & \multicolumn{3}{c}{EN-FR (normal)} & \multicolumn{3}{c}{EN-FR (dense)} & \multicolumn{3}{c}{EN-DE (normal)} & \multicolumn{3}{c}{EN-DE (dense)} \\
			\cmidrule(lr){2-4} \cmidrule(lr){5-7} \cmidrule(lr){8-10} \cmidrule(lr){11-13} & Hits@1 & Hits@10 & MRR & Hits@1 & Hits@10 & MRR & Hits@1 & Hits@10 & MRR & Hits@1 & Hits@10 & MRR \\ 
			\midrule MTransE  & 25.1 & 55.1 & 0.35 & 37.7 & 70.0 & 0.49 & 31.2 & 58.6 & 0.40 & 34.7 & 62.0 & 0.44 \\
			IPTransE & 25.5 & 55.7 & 0.36 & 42.9 & 78.3 & 0.55 & 31.3 & 59.2 & 0.41 & 34.0 & 63.2 & 0.44 \\
			JAPE     & 25.6 & 56.2 & 0.36 & 40.7 & 72.7 & 0.52 & 32.0 & 59.9 & 0.41 & 37.5 & 66.1 & 0.47 \\
			KDCoE    & 22.1 & 47.4 & 0.33 & 54.5 & 85.1 & 0.65 & 34.1 & 56.9 & 0.42 & 58.7 & 79.9 & 0.66 \\ 
			BootEA   & 31.3 & 62.9 & 0.42 & 64.8 & 91.9 & 0.74 & 44.2 & 70.1 & 0.53 & 66.5 & 87.1 & 0.73 \\ 
			\midrule RSN4EA   & \textbf{34.7} & \textbf{63.1} & \textbf{0.44} & \textbf{75.6} & \textbf{92.5} & \textbf{0.82} & \textbf{48.7} & \textbf{72.0} & \textbf{0.57} & \textbf{73.9} & \textbf{89.0} & \textbf{0.79} \\ 
			\bottomrule
	\end{tabular}}
\end{table*}

\subsection{Implementation Details}

We built RSN4EA using TensorFlow. The embeddings and weight matrices were initialized with Xavier initializer, and the embedding size was set to 256. We used two-layer LSTM \cite{LSTM} with Dropout \cite{Dropout}, and conducted batch normalization \cite{BN} for both input and output of an RSN. We used Adam optimizer \cite{Adam} with mini-batch size 512 and learning rate 0.003. We trained an RSN for up to 30 epochs. The random walk biases were set to $\alpha=0.9,\beta=0.9$, and the walk length was set to 15. The source code, datasets and results will be available online.

For the comparative methods, we used the source code provided in their papers except KDCoE, since KDCoE has not released its source code yet. We implemented KDCoE by ourselves. We tried our best effort to adjust the hyper-parameters to make the performance optimal. Following the previous work \cite{JAPE,BootEA}, we used 30\% of reference alignment as prior alignment and chose Hits@1, Hits@10 and mean reciprocal rank (MRR) as evaluation metrics.

\subsection{Results on Entity Alignment}

Tables \ref{tab:mono_results} and \ref{tab:cross_results} depict the EA results on monolingual and cross-lingual datasets, respectively. It is evident that capturing long-term dependencies by paths enables RSN4EA to outperform the existing EA methods.

Generally, the heterogeneity of different KGs is more severe than a KG with different languages. A key module for embedding-based EA methods is to embed the information of entities in different KGs into a unified space. Thus, aligning entities in different KGs is more difficult for embedding-based EA methods. With the help of establishing long-term dependencies, RSN4EA captured richer information of KGs and learned more accurate embeddings, leading to more significant improvement on the more heterogenous datasets (DBP-WD and DBP-YG).

The two tables also demonstrate that the embedding-based EA methods are sensitive to the density. The performance of all the methods on the normal datasets is significantly lower than that on the dense datasets. Although the normal datasets are more difficult, RSN4EA still showed considerable advantages compared with the other methods, since it used long paths to capture implicit connections among entities and represented them in the embeddings.

It is worth noting that RSN4EA showed larger superiority in terms of Hits@1 and MRR. This is due to the fact that Hits@1 only considers the completely correct results, and MRR also favors top-ranked results. As aforementioned, RSN4EA embedded the long-term dependencies into the learned embeddings, which contains richer information to help identify aligned entities in different KGs. The better performance on these two metrics verified this point.

\subsection{Results on KG Completion}
\begin{table}
	\centering
	\caption{KG completion results on FB15K-237}
	\label{tab:link_results}
	{\scriptsize
		\begin{tabular}{lccc}
			\toprule Methods & Hits@1 & Hits@10 & MRR \\ 
			\midrule TransE$^\dagger$ \cite{TransE} & 13.3 & 40.9 & 0.22 \\
			TransR$^\dagger$ \cite{TransR} & 10.9 & 38.2 & 0.20 \\
			\midrule ComplEx \cite{ComplEx}   & 15.2 & 41.9 & 0.24 \\
			NeuralLP \cite{NeuralLP} & --   & 36.2 & 0.24 \\
			ConvE \cite{ConvE}       & \textbf{23.9} & \textbf{49.1} & \textbf{0.31} \\
			\midrule RSN4EA (w/o cross-KG bias)                   & 20.0 & 43.6 & 0.28 \\
			\bottomrule
			\multicolumn{4}{l}{``$\dagger$" denotes the methods executed by ourselves using the provided source} \\
			\multicolumn{4}{l}{\quad \ \,\ code, due to some metrics were not used in literature.} \\
			\multicolumn{4}{l}{``--" denotes the unknown results, due to we cannot obtain the source code.} \\
	\end{tabular}}
\end{table}

Since RSN4EA can train KG embeddings for EA, it is also interesting to apply RSN4EA to KG completion \cite{TransE}, which is one of the most prevalent task for KG representation learning. To achieve this, we removed the cross-KG bias during the random walk sampling and conducted the KG completion experiment. Specifically, for a triple $(s,l,o)$, KG completion aims to predict the object entity $o$ given $(s,l,?)$ or predict the subject entity $s$ given $(?,l,o)$.

FB15K and WN18 are the most widely-used benchmark datasets for KG completion \cite{TransE}. However, recent studies \cite{Node+LinkFeat,ConvE} exposed that these two datasets have the problem of leaking testing data. To solve this issue, a new dataset called FB15K-237 was recommended, and we used this dataset to assess RSN4EA in our experiments. 

The experimental results are shown in Table~\ref{tab:link_results}. ConvE---a method tailored to KG completion---obtained the best results on FB15K-237, followed by our RSN4EA. It is worth noting that, while predicting the entities given one triple is not the primary goal of RSN4EA, it still achieved comparable or better performance than many methods focusing on KG completion, which indicated the potential of leveraging KG paths for learning embeddings.


\section{Further Analysis}
\label{sect:analysis}

\begin{figure}
\centering
\includegraphics[width=\linewidth]{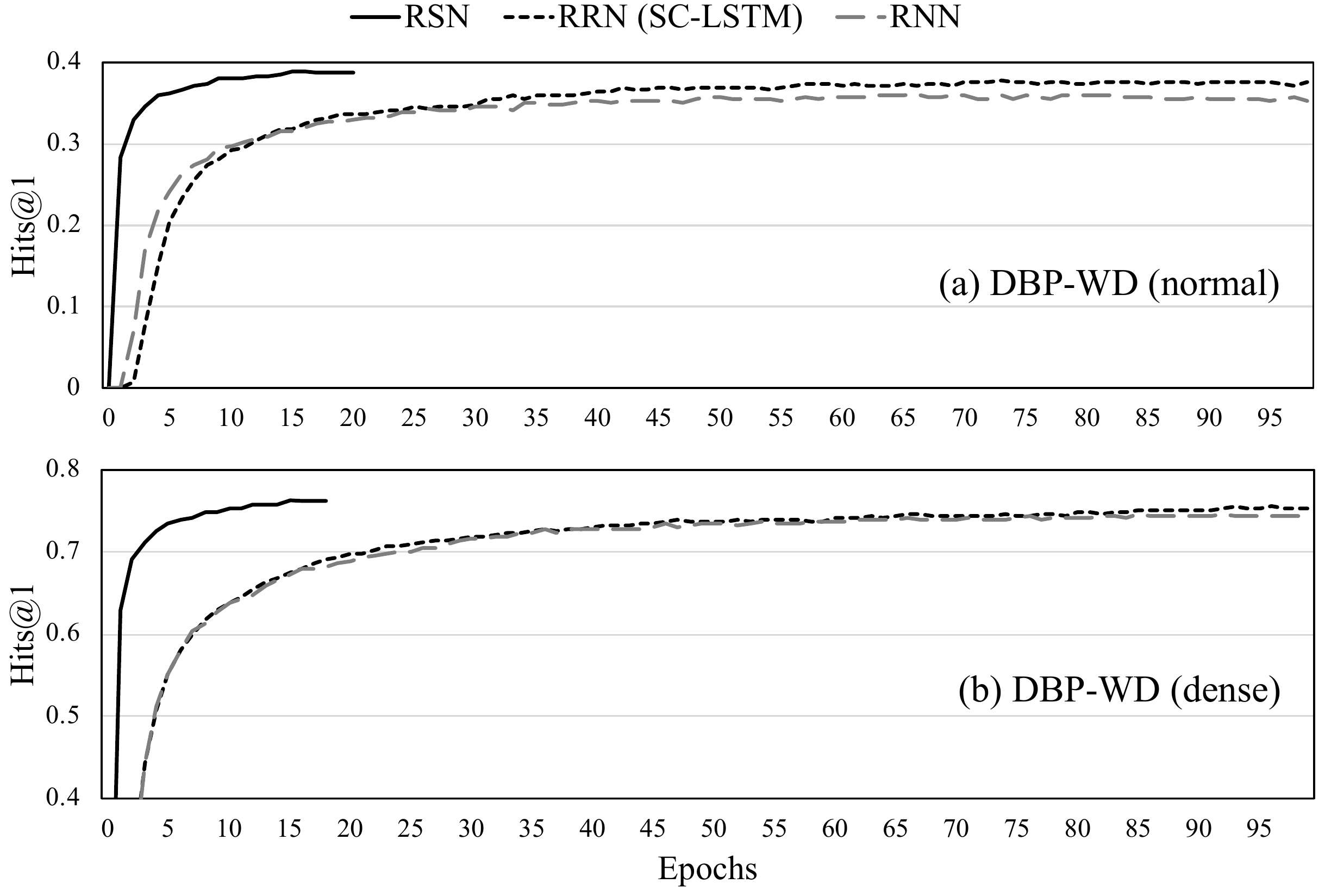}
\caption{Hits@1 results w.r.t. epochs required by alternative networks to converge}
\label{fig:network}
\end{figure}

\subsection{Comparison with Alternative Networks}

To assess the feasibility of RSN, we conducted experiments to compare it with RNN and RRN. Both RNN and RRN were implemented using the same multi-layer LSTM units, Dropout and batch normalization.

The comparison results are shown in Figure~\ref{fig:network}. Since RNN and RRN did not consider the structure of KG paths, they converged the embedding learning at a very slow speed. Compared with RNN, RSN achieved better performance with only $1/30$ time cost, which indicated that this particular residual structure is essential for RSN4EA. Furthermore, RRN is a generic network involving residual learning in the conventional RNN. But it only achieved little improvement compared with RNN. This implied that simply combining residual learning with RNN cannot significantly help KG sequence modeling.

\subsection{Sensitivity to Proportion of Prior Alignment}

The proportion of prior alignment may significantly influence the performance of embedding-based EA methods. However, we may not obtain a large number of prior alignment in practice. We tested the performance of RSN4EA and BootEA (the second best method in our previous experiments) in terms of the proportion of prior alignment from 50\% to 10\% with step 10\%. 

\begin{figure} 
	\centering    
	\subfigure[DBP-WD (normal)]{\includegraphics[width=0.49\columnwidth]{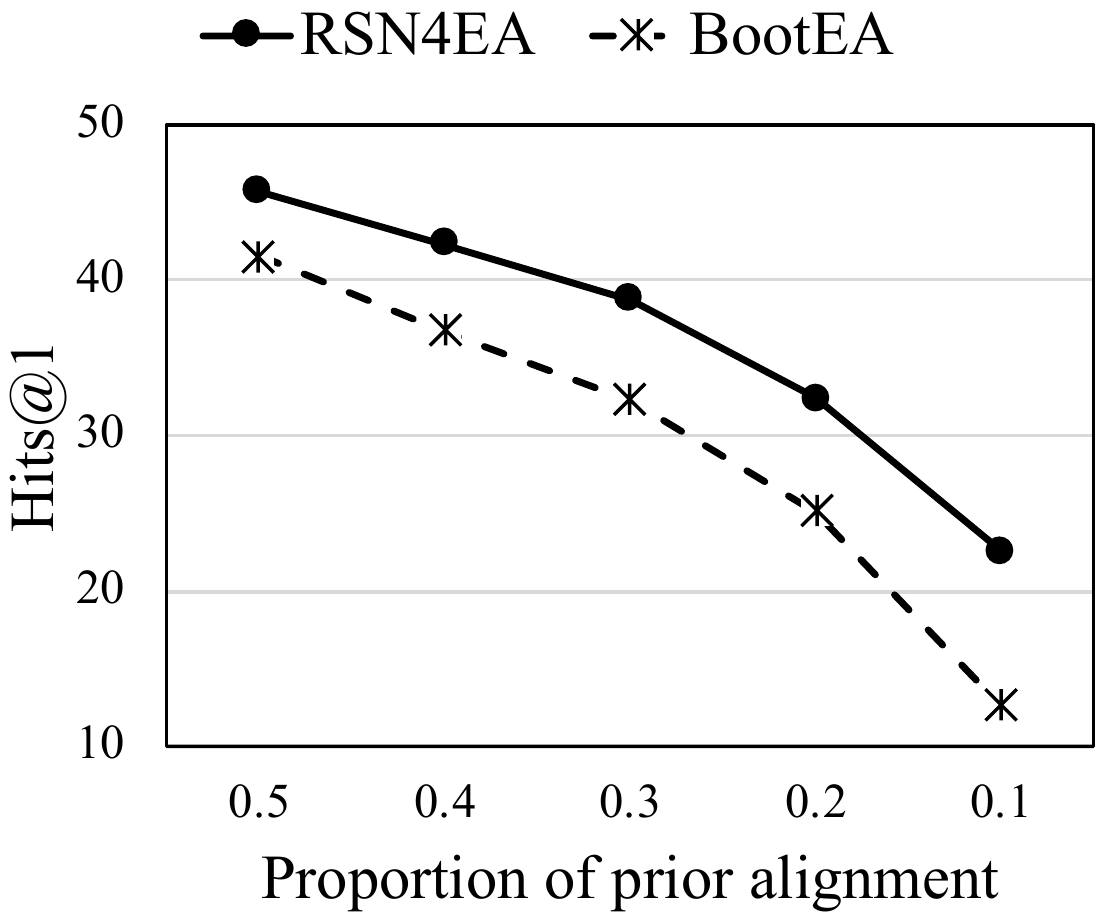}}
	\subfigure[DBP-WD (dense)]{\includegraphics[width=0.49\columnwidth]{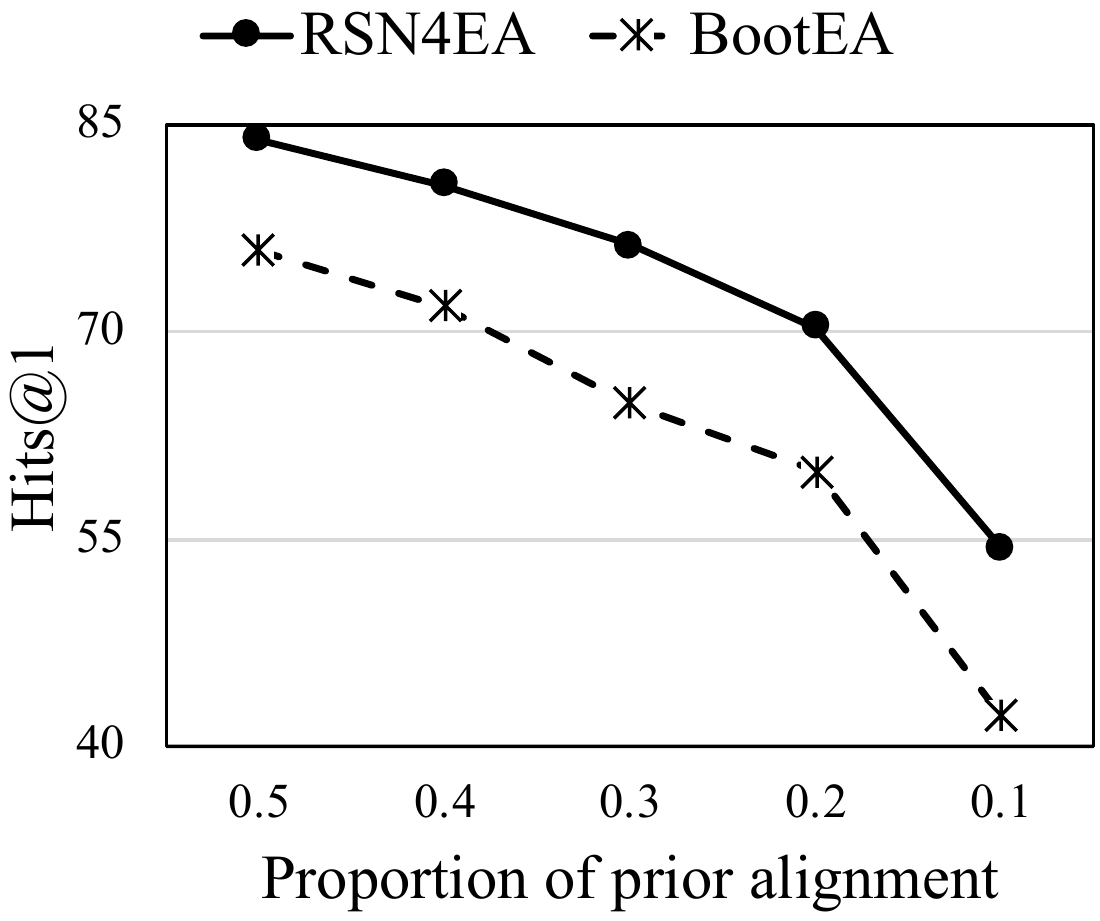}}     
	\caption{Hits@1 results w.r.t. proportion of prior alignment}     
	\label{fig:prior}     
\end{figure}

Due to space limitation, we only depicted the results on the DBP-WD dataset in Figure~\ref{fig:prior}. The performance of the two methods continually dropped with the decreasing proportion of prior alignment. However, the curves of RSN4EA are gentler than BootEA. Specifically, on the normal dataset, for the four proportion intervals, RSN4EA lost 7.4\%, 8.2\%, 16.5\% and 30.2\% on Hits@1 respectively, while BootEA lost 11.8\%, 12.0\%, 22.3\% and 49.8\% respectively, which demonstrated that RSN4EA is a more stable method. Additionally, when the proportion was down to 10\%, the Hits@1 result of RSN4EA on the normal dataset was almost twice higher than that of BootEA, which indicated that modeling paths helps RSN4EA propagate the identity information across KGs more effectively and alleviates the dependence on the proportion of prior alignment.

\subsection{Sensitivity to Random Walk Length} 

\begin{figure} 
	\centering    
	\subfigure[Normal datasets]{\includegraphics[width=0.49\columnwidth]{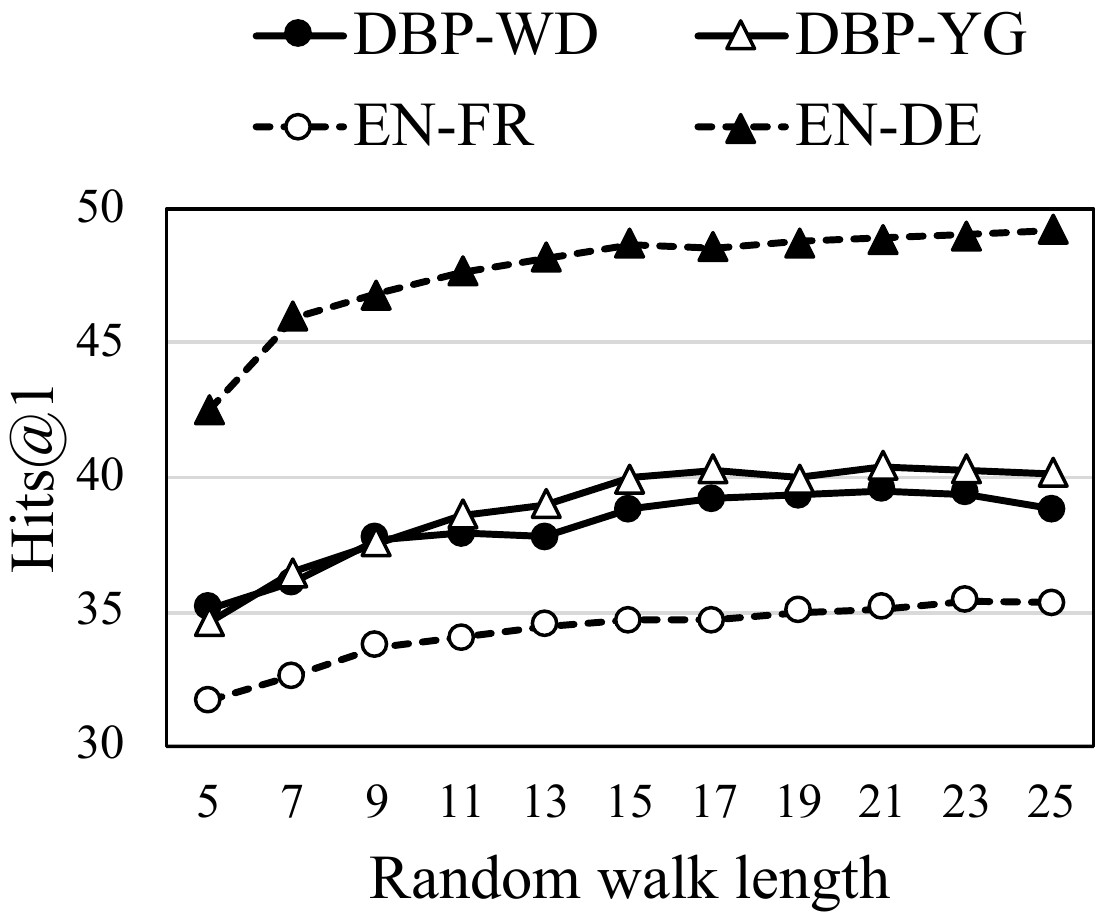}}
	\subfigure[Dense datasets]{\includegraphics[width=0.49\columnwidth]{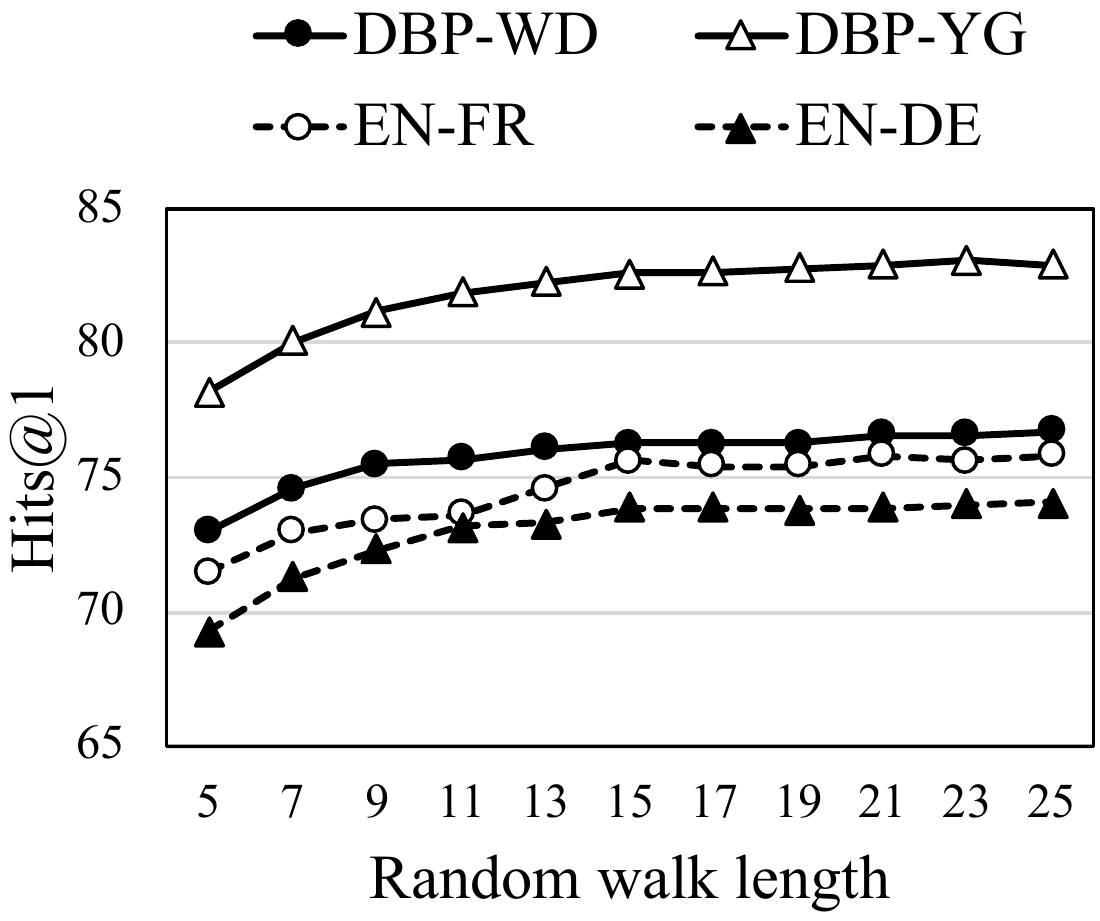}}     
	\caption{Hits@1 results w.r.t. random walk length}     
	\label{fig:walk}     
\end{figure}

We also observed how the random walk length affected the EA performance. As shown in Figure~\ref{fig:walk}, on all the eight datasets, the Hits@1 results increased sharply during length 5 to 15, which indicates that modeling longer paths can help learn KG embeddings and obtain better performance. Furthermore, we observed that the performance approached to saturation for length 15 to 25. Therefore, in consideration of the efficiency, the results reported in Tables \ref{tab:mono_results} and \ref{tab:cross_results} are based on length 15.

\section{Conclusion and Future Work}
\label{sect:concl}
In this paper, we proposed RSN4EA, which employs biased random walks to sample paths specific to EA, and leverages RSN for learning KG embeddings. Our experimental results showed that RSN4EA not only outperformed the existing embedding-based EA methods, but also achieved superior performance compared with RNN and RRN. It also worked well for KG completion.

In future work, we plan to continue exploring KG sequence learning. First, KGs often contain rich textual information like names and descriptions. Such information can be modeled with character-/word-level sequential models. RSN is capable of modeling KGs in a sequential manner, therefore it is worth studying a unified sequential model to learn KG embeddings using all valuable information. Second, in addition to paths, the neighboring information provides another type of context and may be also helpful for learning KG embeddings. We look forward to integrating the neighboring context to further improve the performance.


\bibliographystyle{aaai}
\bibliography{references}

\end{document}